# Toward Abstraction from Multi-modal Data: Empirical Studies on Multiple Time-scale Recurrent Models

Junpei Zhong*[†][‡], Angelo Cangelosi[†] and Tetsuya Ogata*[‡]
*National Institute of Advanced Industrial Science and Technology (AIST), Aomi 2-3-26, Tokyo, Japan
Email: zhong@junpei.eu
[†]Centre for Robotics and Neural Systems, Plymouth University, Plymouth, UK
[‡]Lab for Intelligent Dynamics and Representation, Waseda University, Tokyo, Japan

*Abstract*—The abstraction tasks are challenging for multi-modal sequences as they require a deeper semantic understanding and a novel text generation for the data. Although the recurrent neural networks (RNN) can be used to model the context of the time-sequences, in most cases the long-term dependencies of multi-modal data make the back-propagation through time training of RNN tend to vanish in the time domain. Recently, inspired from Multiple Time-scale Recurrent Neural Network (MTRNN) [1], an extension of Gated Recurrent Unit (GRU), called Multiple Time-scale Gated Recurrent Unit (MTGRU), has been proposed [2] to learn the long-term dependencies in natural language processing. Particularly it is also able to accomplish the abstraction task for paragraphs given that the time constants are well defined. In this paper, we compare the MTRNN and MTGRU in terms of its learning performances as well as their abstraction representation on higher level (with a slower neural activation). This was done by conducting two studies based on a smaller data-set (two-dimension time sequences from non-linear functions) and a relatively large data-set (43-dimension time sequences from iCub manipulation tasks with multi-modal data). We conclude that gated recurrent mechanisms may be necessary for learning long-term dependencies in large dimension multi-modal data-sets (e.g. learning of robot manipulation), even when natural language commands was not involved. But for smaller learning tasks with simple time-sequences, generic version of recurrent models, such as MTRNN, were sufficient to accomplish the abstraction task.

## I. INTRODUCTION

The long-term dependencies of the natural language sentences are difficult to be learnt [3] by Vanilla recurrent network because in most cases the gradients tend to vanish in time while the back-propagation through time is being processed [4, 5]. This makes most gradient-based learning methods for recurrent neural networks hardly form a long-term effect. To solve this problem, the earliest attempt was the long short-term memory (LSTM) [6] which consists of various gating functions that controlled by simple element-wise operations. Since it was designed, it has achieved satisfaction results in competitions [7] as well as tasks such as dialogue system [8], sentiment analysis [9] and machine translation [10].

The Gated Recurrent Unit (GRU) models [11], as a more efficient version than the LSTM [12], has recently been used widely for language processing where they were also able achieve state-of-the-art results with less computation requirements than LSTM, as a GRU has fewer control gates than the LSTM unit does [12]. Despite of the differences in their internal operations, both of them can efficiently eliminate the gradient vanishing problem with the following common features:

- They can store the previous activations in the internal memories which can be later refreshed or be retrieved, depending on different contexts;
- These operations on the internal activations are controlled by different gates within the recurrent units.
- The control policies of the gates are learnt by the context of the training sequences; They form the composition operations which control the information flow that goes in and out of the internal memory.

With all these features, the recurrent units have the ability to modify their internal weights (i.e. internal structures) based on the long-term dependencies existing in the temporal sequences to the cell states, given that the gated structures are well-trained. Furthermore, while input and/or output are of variable length, the gated-like units stack in a hierarchical manner [13] are also able to extract the neural representation with a fixed length, based on the time dependencies in the temporal domain, in which the *unpredictable* inputs of lower level of RNN become inputs to the connecting higher level units, where a slower activation is updated [14]. This is also one of the theoretical foundations of the state-of-the-art deep learning methods. In the context of language processing, the higher level units of deep learning architectures can represent the extracted meaning of the phrase/sentence, while the inputs are (almost) raw data with one-hot/embedding word representations of this phrase/sentence. Furthermore, connecting and training with two deep (recurrent or convolutional) networks with a shared higher level representation, namely encoder-decoder architecture, can abstract meanings for the sentences, even if such "sentences" are in different languages or modalities. As a result, a few applications such as image captioning [15] (LSTM + CNN) and machine translation [10] (two LSTMs) have been developed based on such encoder-decoder architecture.

While previous architectures used the encoder-decoder architecture to connect visual images and language sequences, in this paper, we proposed that it is possible to do abstraction tasks for multi-modal information when applying the hierarchical RNN architectures, especially with gated-like units, to the sensorimotor information sequences obtained from robotic platforms and language sequences. This is an extension work of our previous experiments [16] based on the Multiple Timescale Neural Network (MTRNN) [1]. Inspired by the time-constant concept of MTRNN, a Multiple Time-scale Gated Recurrent Units (MTGRU) were recently proposed [2] to apply this idea into gated-like recurrent units to accomplish the text extraction task. Moreover, its dynamic representation on higher levels along time makes it an ideal architecture to connect natural language commands, the dynamic multi-modal environment and the motor actions for robotic systems. Therefore, we herein conducted experiments about robot manipulation based on the MTGRU network. We also did comparison about the performances between MTRNN and MTGRU. The organization of this paper is as follows: a brief introduction of MTRNN and MTGRU model is presented at the next section. The empirical studies for algorithmic and multi-modal data from iCub manipulation are showed at the third section. At the last section, discussion and summaries will be given.

## II. MODELS

A recurrent neural network (RNN) is a feed-forward neural network with directed connecting weights. As the weights form a directed connection between neural units in the time-domain, a neural unit that is connected with the recurrent weights is dependent on neural activity at the previous time-step(s). With sufficient learning, it is able to model a variable-length sequence input data.

More formally, given a sequence $x = (x_1, x_2, , x_t)$, the RNN updates its recurrent hidden state $h_t$ by Eq.1:

$$h_t = \begin{cases} 0, t = 0 & (1a) \\ \Phi(h_{t-1}, x_t), t > 0 & (1b) \end{cases}$$

where $\Phi$ is a non-linear function. Ideally, given the hidden states of the network, the output $y = (y_1, y_2, \cdots, y_t)$ is computed as

$$p(y_1, y_1, \cdots, y_t) = p(y_1) \cdot p(y_2|x_1) \cdot p(y_3|x_1, x_2) \cdots$$
$$p(y_t|x_1, x_2, \cdots, x_{t-1}) \quad (2)$$

In the case of RNN, the last term $p(y_t|x_1, x_2, \cdots, x_{t-1})$ can be presented as the activation of hidden unit at time $t$:

$$p(y_t|x_1, x_2, \cdots, x_{t-1}) = g(h(t)) \quad (3)$$

where $h(t)$ is from Eq. 1. And the term $h(t)$ is the units we investigated in the empirical studies below, in which we could observe the abstract information from previous time-steps.

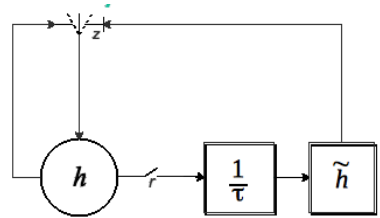

Fig. 1: The MTGRU Unit

### A. Multiple Time-scale Recurrent Neural Network

In the MTRNN network [1], the learning of each neuron follows the updating rule of classical firing rate models, in which the activity of a neuron is determined by the average firing rate of all the connected neurons. Additionally, the neuronal activity is also decaying over time following an updating rule of leaky integrator model.

Assuming the $i$-th MTRNN neuron has the number of $N$ connections, the current membrane potential status of a neuron can be defined as both by the previous activation as well as the current synaptic inputs:

$$u_{i,t+1} = (1 - \frac{1}{\tau_i})u_{i,t} + \frac{1}{\tau_i}[\sum_{j \in N} w_{i,j} x_{j,t}] \quad \text{(if } t > 0) \quad (4)$$

where $w_{i,j}$ represents the synaptic weight from the $j$-th neuron to the $i$-th neuron, $x_{j,t}$ is the activity of $j$-th neuron at $t$-th time-step and $\tau$ is the time-scale parameter which determines the decay rate of this neuron: a larger $\tau$ means their activities change slowly over time compared with those with a smaller $\tau$. As we can see, the MTRNN essentially is a continuous recurrent model. Therefore, it also exists vanished gradient problem.

### B. Multiple Time-scale Gated Recurrent Units

Although both GRU and LSTM have gating mechanisms for the recurrent units, compared with the three gates that exist in LSTM, a GRU has only two gates: a reset gate $r$ and an update gate $z$. As the names imply, the reset gate determines how to combine the current input with the previous status of internal memory, and the update gate defines how much of the previous memory to be preserved. The basic idea of using such a gating mechanism to learn long-term dependencies is similar as in a LSTM, but it was reported that fewer number of gates leads to more efficient in training [12].

When the concept of multiple time-scales (MT) is applied in GRU, it has a similar meaning as in MTRNN: it summarises the dynamics with different time scales of the temporal sequences. Compared with GRU, the output of the multiple time-scale gated recurrent units (MTGRU) contains a so-called "time-scale" constant, which controls how the output from previous time steps influences the current output. It is equivalent that this constant is being multiplied to the output and modulates the mixture of the current and previous states.

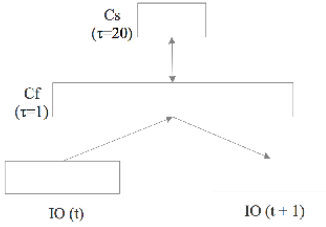

Fig. 2: The Same Network Architecture was chosen for both MTGRU and MTRNN

In Fig. 1, the internal structure of MTGRU is shown, which demonstrates how the candidate activation $\tilde{h}$ is multiplied with constant $1/\tau$ to the current output. In the mean while, the reset gate $r_t$, update gate $z_t$, and the candidate activation $u_t$ are computed similarly to those of the original GRU in [11].

$$r_t = \sigma(W_{xr}x_t + W_{hr}h_{t-1}) \quad (5)$$
$$z_t = \sigma(W_{xz}x_t + W_{hz}h_{t-1}) \quad (6)$$
$$u_t = tanh(W_{xu}x_t + W_{hu}(r_t \odot h_{t-1})) \quad (7)$$
$$h_t = ((1-z_t)h_{t-1} + z_t u_t)\frac{1}{\tau} + (1-\frac{1}{\tau})h_{t-1} \quad (8)$$

Similar as the MTRNN, the pre-defined time-scale $\tau$ is introduced to the activation term $h_t$ at Eq. 8 to control the levels abstraction. The time-constant controls in what ratio the current and past output to the GRU cell are mixed to compute. A larger $\tau$ indicates the past activations have larger influences to the current activation, presenting the long term dynamic feature of the temporal sequences.

In the original MTGRU paper [2], the learning formula of the MTGRU and the performances of MTGRU in abstraction was presented. In this paper, we concentrate the abstraction of multi-modal sequences obtained from a humanoid robot, especially its difference with MTRNN. Starting from simple sequences, we conducted two empirical studies based on the MTGRU units.

### III. EMPIRICAL STUDIES

In this section, we conducted two case studies on a simple time-sequence-learning task and a more complicated multi-modal-sequence learning task. In order to have a fair comparison, the same architecture with the same parameters were used for both MTRNN and MTGRU. As shown in Fig. 2, the architecture for these empirical studies contains three layers: an input-output layer ($IO$) and two context layers called Context fast ($C_f$) and Context slow ($C_s$). The values of the time constants here were obtained with experiments in [16]. The Input-output neurons have full connections with the fast context layers. And the slow context layer only connects with the fast context layer, representing a slower feature that extracts from the fast context dynamics.

In the following text, we denote the indices of these neurons as:

$$I_{all} = I_{IO} \cup I_{C_f} \cup I_{C_s} \quad (9)$$

where $I_{IO}$ represents the indices to the neurons at the input-output layer, $I_{C_f}$ belongs to the neurons at the context fast layer and $I_{C_s}$ belongs to the neurons at the context slow layer. We adopted a tanh function on the $IO$ layer, then the corresponding RNN functions on context layers:

$$x_{cf} = \tanh_k(i_t), k \in I_{IO} \quad (10)$$
$$x_{cs} = y_{cf} = RNN_k(x_{cf}), k \in I_{C_f} \quad (11)$$
$$y_{cs} = RNN_k(x_{cs}), k \in I_{C_s} \quad (12)$$
$$i_{t+1} = o_t = RNN_k(x_{cf} + y_{cs}), k \in I_{C_f} \quad (13)$$

where RNN functions represent either MTRNN or MTGRU functions in the $k$th neuron. Note that in MTRNN, the neurons on one layer own full connectivity to all neurons within the same and adjacent layers. In MTGRU, as we introduced before, the internal activations also have full connectivities with inputs and outputs. Therefore, the only difference between two architectures of MTRNN and MTGRU are the neural activations within each neuron units.

While training multiple sequences, both the MTRNN and MTGRU should balance the training epochs for each sequence and over-fitting should be avoided. Therefore, one epoch was defined to include a few iterations for training all the sequence using stochastic gradient descent (SGD) [17, 18], as showed in Algorithm 1:

---
**Algorithm 1** Multiple Time-sequences Training
---
1: **procedure** ONE EPOCH($data$) ▷ $data$ contains multiple time sequences.
2:    **for** $seq \in data$ **do**
3:       **while** $error > threshold$ or $iteration > maximum\_iteration$ **do**
4:       ▷ Repeat iteration for one sequence until threshold is achieved
5:          Run SGD($seq$)
6:       **end while**
7:    **end for** ▷ Choose the next sequence
8: **end procedure**
---

#### A. Case 1: Simple Non-linear Sequences Abstraction

In this case study, two time sequences that include two dimensions were generated to examine the learning performances of MTRNN and MTGRU. The two dimensions X = $[x_1, x_2]$ of the first sequence was defined as:

$$X_1 = \begin{cases} x_1 = sin2t & (14a) \\ x_2 = sint & (14b) \end{cases}$$

And the second sequence was defined as:

$$X_2 = \begin{cases} x_1 = sin2t \cdot cos3t & (15a) \\ x_2 = \dfrac{sin3t}{2t} \cdot \dfrac{sint}{t} - 0.5 & (15b) \end{cases}$$

In both cases, 100 time-steps were applied, i.e. $t = \dfrac{k-50}{50} \cdot \pi, k = \{0, 1, 2, \cdots, 99\}$.

The parameters in the MTRNN and MTGRU experiments are shown as Tab. I. The parameters $n_{C_f}$ and $n_{C_s}$ in the case of MTRNN mean the numbers of neurons on $C_f$ and $C_s$ layers, while in the case of MTGRU, they mean the number of dimensions of the MTGRU unit on the $C_f$ and $C_s$ layers.

Note that, compared with our previous experiment [16], we did not employ the SOM pre-processing because a fair comparison between MTRNN and MTGRU is needed.

TABLE I: MTRNN & MTGRU Parameters (Case 1)

| Parameters | Parameter's Descriptions | Value |
|---|---|---|
| $\eta$ | Learning Rate | $10^{-4}$ |
| $n_{C_f}$ | Size of $C_f$ | 100 |
| $n_{C_s}$ | Size of $C_s$ | 5 |
| $\tau_f$ | Time-constant of $C_f$ | 1 |
| $\tau_s$ | Time-constant of $C_s$ | 20 |
| $max\_iteration$ | Max. iteration for training one sequence | 2000 |
| $threshold$ | Threshold for early stop | $10^{-3}$ |
| $\alpha$ | Mixed ratio for prediction/real | 0.9 |

With the network implemented in Theano [19], the training was done on an AWS G2 (2x large) server equipped with Grid K520. The training curve of MTRNN and the MTGRU were depicted in Figs. 3. We can see that with the same learning rate, the error of the MTRNN converged faster than the MTGRU, which was compatible with our intuition that MTGRU converges slower because it owns more weights than the MTRNN, although they have the same parameters.

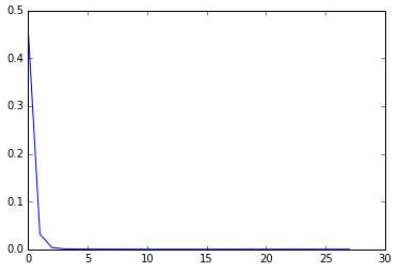

(a) Training Curve with MTRNN (Case 1)

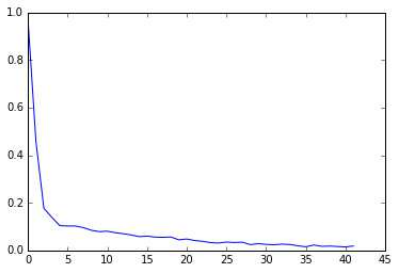

(b) Training Curve of MTGRU (Case 1)

Fig. 3: Training Curves with MTRNN and MTGRU

The quantitative results about the performances of MTRNN and MTGRU were shown in Tab. II. And the comparison between the MTRNN and MTGRU output and the real value of Seq.1 were shown in Fig. 6.

TABLE II: MTRNN & MTGRU Performances (Case 1)

| | MTRNN | MTGRU |
|---|---|---|
| Prediction Error (RMS after 30 Epochs) | 1.6353 | 2.2255 |
| Time per GD (ms) | 135.66 | 226.26 |

To further examine the internal dynamics of both networks, we selected the neural activities of the $C_f$ and $C_s$ layers while the Seq. 1 ($X_1$) was as the input. From the internal dynamics of the context layers, we could observe significant difference between the dynamics of MTRNN (Fig. 4) and MTGRU (Fig. 5):

- oscillations in the activation could be found in the MT-GRU context units.
- the range of neural dynamics in MTRNN was significantly larger than in MTGRU.

### B. Case 2: Multi-modal Data Abstraction

To examine the network performance in more complicated tasks such as abstraction from robot multi-modal data, we recorded the multi-modal data from object manipulation experiments based on an iCub robot [20]. iCub is a child sized humanoid robot which was built as a testing platform for theories and models of cognitive science and neuroscience. Mimicking a two-year old infant, this unique robotic platform has 53 degrees of freedom totally. As such, using the iCub, we set a learning scenario in which a human instructor was teaching the robotic learner a set of language commands whilst providing kinaesthetic demonstration of the named actions as well as the corresponding visual inputs from the camera. The target of this case study was to evaluate the performances of MTRNN and MTGRU in this complicated task with a large data-set, which may toward a natural language understanding for humanoid robots.

TABLE III: Dictionaries of verbs and nouns for the data sets: The instructor showed the robot with different combinations from the 9 actions and 9 nouns. The actions and the objects are represented in two discretised values for semantic command inputs which range from $0 - 0.9$. For instance, the command "lift [the] ball" is translated into values $[0.8, 0.2]$.

| Actions | Slide Left | Slide Right | Touch | Reach | Push |
|---|---|---|---|---|---|
| Verb Value | 0.0 | 0.1 | 0.2 | 0.3 | 0.4 |
| Actions | Pull | Point | Grasp | Lift | |
| Verb Values | 0.5 | 0.6 | 0.7 | 0.8 | |
| Objects | Tractor | Hammer | Ball | Bus | Modi |
| Noun Value | 0.0 | 0.1 | 0.2 | 0.3 | 0.4 |
| Objects | Car | Cup | Cubes | Spiky | |
| Noun Values | 0.5 | 0.6 | 0.7 | 0.8 | |

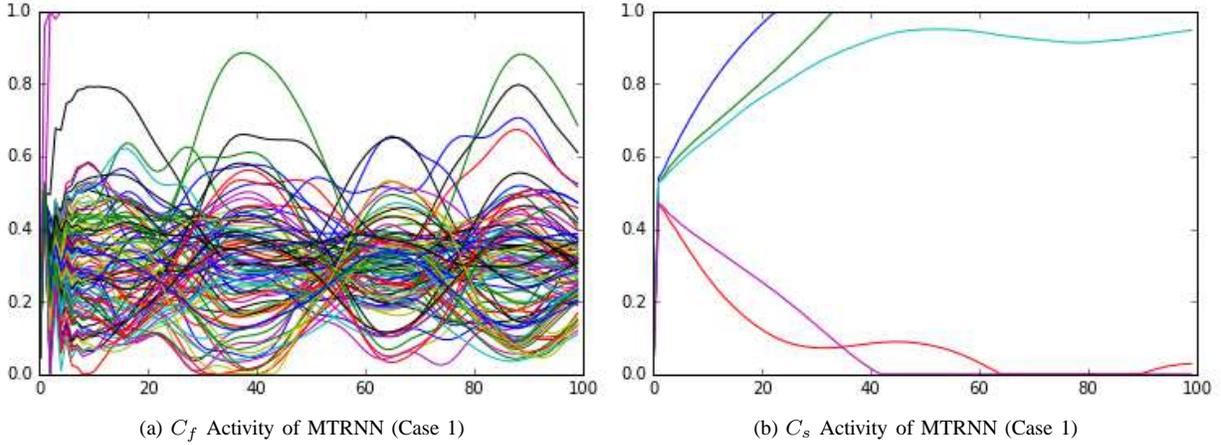

(a) $C_f$ Activity of MTRNN (Case 1)  (b) $C_s$ Activity of MTRNN (Case 1)

Fig. 4: Neural Activity of MTRNN (Case 1)

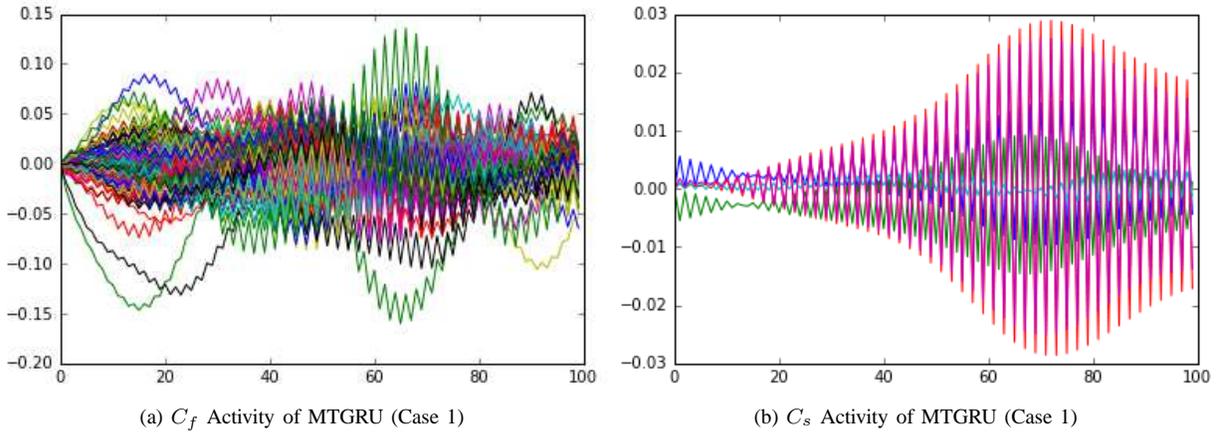

(a) $C_f$ Activity of MTGRU (Case 1)  (b) $C_s$ Activity of MTGRU (Case 1)

Fig. 5: Neural Activity of MTGRU (Case 1)

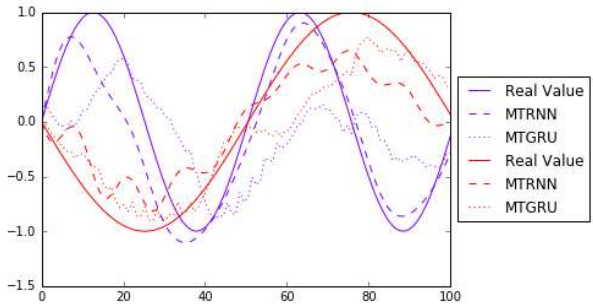

Fig. 6: Predicted and real values of MTRNN and MTGRU (Seq. 1)

*1) Experimental Setup:* Fig. 7 shows the setup used in our manipulation experiments to collect the multi-modal data-set. It was obtained using the following steps:

1) The 9 different objects with significantly different colours and shapes were placed at 6 different locations along a line on the table in front of the iCub.
2) A vocal command was spoken by an instructor according to the visual scene that was perceived by the iCub. A complete combination in a sentence of the vocal command is composed of a verb and a noun. The corresponding verb and noun were recognised and then translated into two dedicated discrete values based on the verb and noun dictionaries like we did in the previous experiment in [16] (Tab. III)[1].
3) The built-in vision tracker of the iCub searched for a ball-shaped object based on the dictionary-generated values using its vision tracker system.
4) Once the object was located, the iCub rotated its head and triggered the object tracking, which changed the encoder values of the neck and eyes.

[1]The speech recognition is not always successful here. In theory it is the same if we use transcribed language data. But toward more natural training process with human instructors, we manually monitored the recognition results before the data sequences were recorded to ensure a better performance. Please also see http://www.nuance.co.uk/dragon/index.htm.

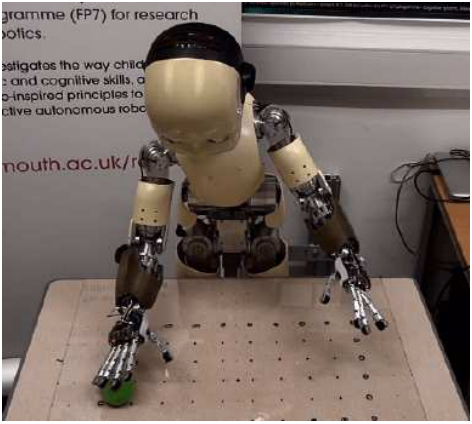

Fig. 7: Data Collection from iCub Robot

5) Joint positions of the head and neck were recorded. The sequence recorder module of the iCub was used to record the sensorimotor trajectories while the instructor was guiding the robot by holding its arms to perform a certain action for each object.

The whole experimental data-set for the iCub manipulation included combinations of 9 actions and 9 objects. In each of the 43-dimensional temporal sequence, it includes the vocal commands (i.e. a complete sentence includes verb and noun), the visual information (presented as joint angles of neck and eyes) and the change of torso angles (resulting in motor actions) sequences. We used such a large number of data to test how the MTRNN and MTGRU perform in such a complicated task. We also aimed at applying the power of recurrent network [21] in natural language processing in robotic platforms, especially to apply humanoid robots in cognitive tasks such as multi-modal interaction and dialogue robots.

*2) Experiment Results:* In this case study, the parameters of both models were also kept the same as shown in Tab. IV.

TABLE IV: MTRNN & MTGRU Parameters (Case 2)

| Parameters | Parameter's Descriptions | Value |
|---|---|---|
| $\eta$ | Learning Rate | $10^{-4}$ |
| $n_{C_f}$ | Size of $C_f$ | 450 |
| $n_{C_s}$ | Size of $C_s$ | 8 |
| $\tau_f$ | Time-constant of $C_f$ | 1 |
| $\tau_s$ | Time-constant of $C_s$ | 20 |
| $max\_iteration$ | Max. iteration for training one sequence | 5000 |
| $threshold$ | Threshold for early stop | $10^{-3}$ |
| $\alpha$ | Mixed ratio for prediction/real | 0.9 |

Figs. 8 show the training curves from MTRNN and MTGRU respectively. Quite different from the previous study, although the MTGRU converged slower than MTRNN, it converged in a more steady way. The training of MTRNN, on the other hand, often converged to local minimum and took more epochs to the same level of error of MTGRU. An output example of the 50th sequence in the data-set were shown in Fig. 9. The quantitative results of the training can be found in Tab. V. As we expected, the training of MTGRU also took almost twice of the computational time in one iteration compared with MTRNN. One interesting thing was that the computation time for one iteration in this case is less than in the previous case, which was probably because of the advantage of GPU for parallel computing in neural networks.

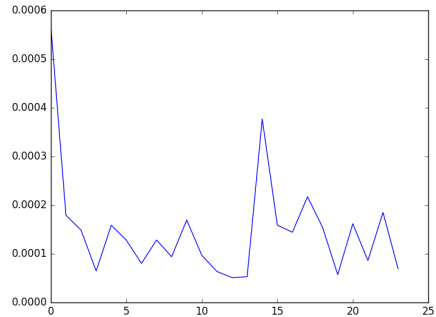

(a) Training Curve with MTRNN (Case 2)

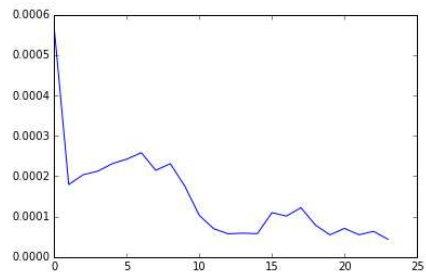

(b) Training Curve with MTGRU (Case 2)

Fig. 8: Training Curves with MTRNN and MTGRU

In Figs 10 and Figs 11, the internal dynamics of the context neurons were also depicted. But since the number of dimensions was too large to examine in details, we made a PCA to reduce the dimension to 2 before demonstrating the neural dynamics, in which we could observe that in the case of larger dimensions, oscillation in dynamics can also be found in MTGRU.

TABLE V: MTRNN & MTGRU Performances (Case 2)

|  | MTRNN | MTGRU |
|---|---|---|
| Prediction Error (RMS after 30 Epochs) | 1.5212 | 1.3677 |
| Time per GD (ms) | 31.79 | 62.54 |

## IV. DISCUSSIONS AND CONCLUSIONS

Inspired by the multiple time-scales which determine the updating rate for the membrane activities in continuous neurons as MTRNN did, the MTGRU was recently proposed as an extended version of GRU model. In this paper, the empirical studies comparing with MTRNN and MTGRU were conducted in terms of its training performances and their feasibility in abstraction for time sequences. Specifically, two cases were studied: 1) the 2-dimensional non-linear time sequences (Sec. III-A); 2) the 43-dimensional multi-modal time sequences (Sec. III-B). As expected, with the two data-sets we provided,

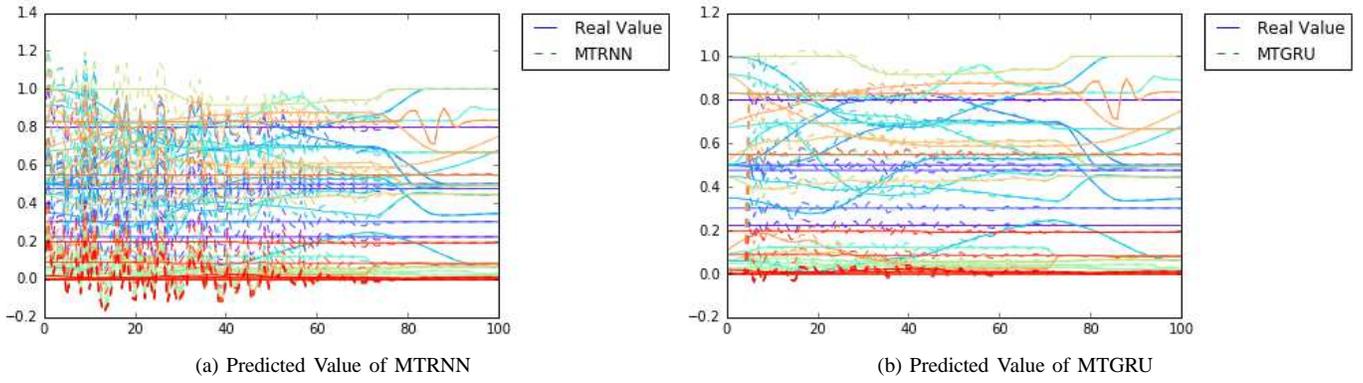

(a) Predicted Value of MTRNN

(b) Predicted Value of MTGRU

Fig. 9: Predicted and real value of MTRNN and MTGRU (50th Seq.)

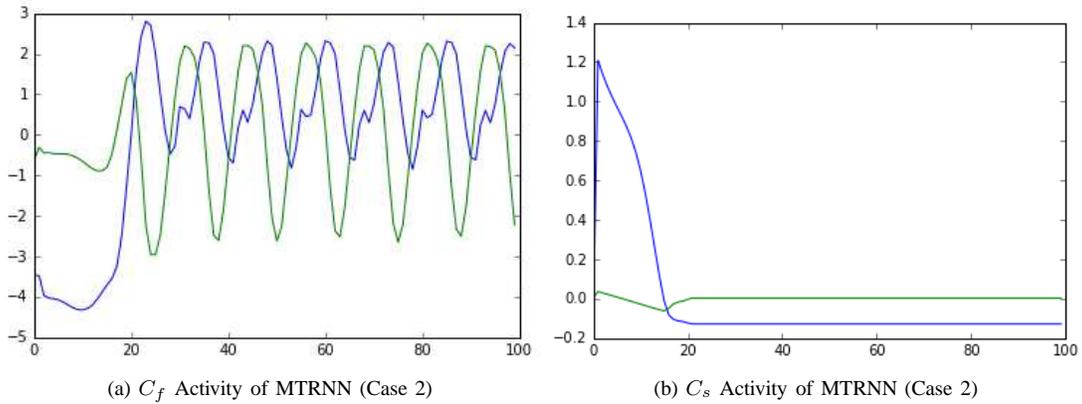

(a) $C_f$ Activity of MTRNN (Case 2)

(b) $C_s$ Activity of MTRNN (Case 2)

Fig. 10: Neural Activity of MTRNN (50th Seq, Case 2)

the complexity of training in GRU (i.e. gates inside the units) costed more computational effort than the MTRNN. We could conclude that for such relatively trivial tasks (without significant long-term dependencies in sequences), the advantage of GRU (possibly LSTM as well) was hardly been exhibited.

However, we also noticed the training of MTGRU for large dimension data converges faster than MTRNN (case 2). This was probably because that the robot manipulation data we used actually exhibits long-term dependencies to some extent. For instance, the movement of hands for grasping depends on the verb showed in the command sentence. If we use more sophisticated time-dependency data in the multi-modal experiments, the gated mechanisms may result in more steady training performance than ordinary RNNs. Furthermore, according to the previous literature with natural language modelling, the gated mechanisms RNN would be necessary to model the long-term dependencies in multi-modal environment when the language commands are involved.

In future work, we will further investigate the following two topics:

- We will investigate the internal dynamics of MTGRU; for example, how the neural oscillation on the context layers happened is still unknown;

- We plan to use natural language as robot commands while using word2vec [22] as a pre-processed input as [23] did, instead of the look-up table III. The final target of this work is to apply multi-modal understanding for both sensorimotor and language temporal sequences on robotic systems.

## APPENDIX

The code of MTGRU can be found on Github[2].

## ACKNOWLEDGMENT

The research was supported by Waseda SGU Program, the EU project POETICON++ under grant agreement 288382 and the UK EPSRC project BABEL. JZ would like to thank FH for the working space provided when the paper was being drafted.

[2]https://github.com/jonizhong/mtgru.git

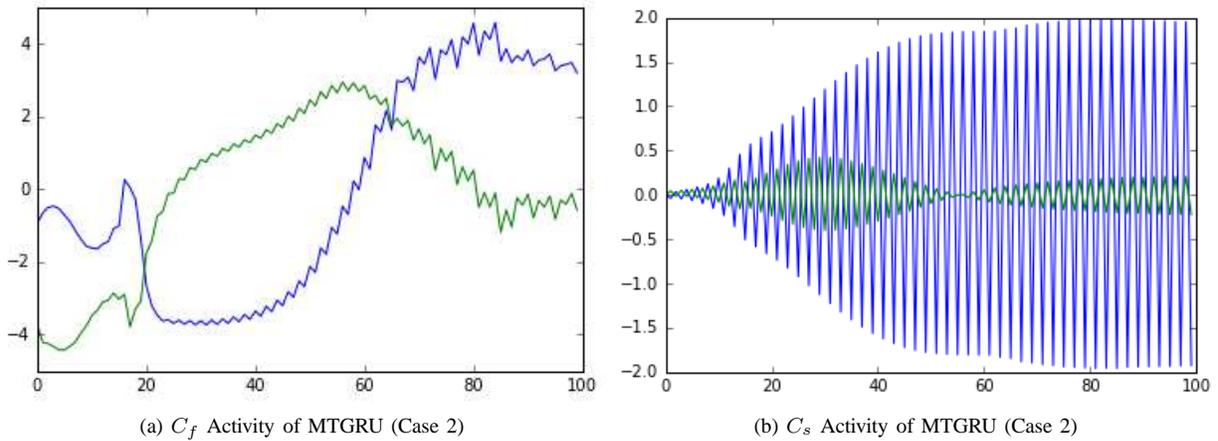

(a) $C_f$ Activity of MTGRU (Case 2)  (b) $C_s$ Activity of MTGRU (Case 2)

Fig. 11: Neural Activity of MTGRU (50th Seq, Case 2)